\documentclass[10pt,twocolumn,letterpaper]{article}

\usepackage{iccv}
\usepackage{times}
\usepackage{epsfig}
\usepackage{graphicx}
\usepackage{amsmath}
\usepackage{amssymb}

\newcommand{\reals}[1]{\mathbb{R}^{#1}}
\newcommand{\set}[1]{\left\{{#1}\right\}}
\newcommand{\transpose}[1]{#1^\top}

\newcommand{\comment}[1]{}

\usepackage{pifont}
\newcommand{\cmark}{\ding{51}}%
\newcommand{\xmark}{\ding{55}}%
\usepackage{multirow}
\usepackage{subcaption} 
\usepackage{caption}
\usepackage{multicol}

\usepackage[pagebackref=true,breaklinks=true,letterpaper=true,colorlinks,bookmarks=false]{hyperref}

\iccvfinalcopy 

\def\iccvPaperID{****} 

\begin{document}

\title{Video Representation Learning by Dense Predictive Coding}

\author{Tengda Han
	\and
	Weidi Xie
	\and
	Andrew Zisserman
	\and 
	Visual Geometry Group, Department  of Engineering Science, University of Oxford\\
	{\tt\small \{htd, weidi, az\}@robots.ox.ac.uk}
}

\twocolumn[{%
	\renewcommand\twocolumn[1][]{#1}%
	\maketitle
	\vspace{-12mm}
	\setlength{\tabcolsep}{2pt}
	\begin{center}
		\centering
		\begin{tabular}{c}
			\includegraphics[width=1.0\textwidth]{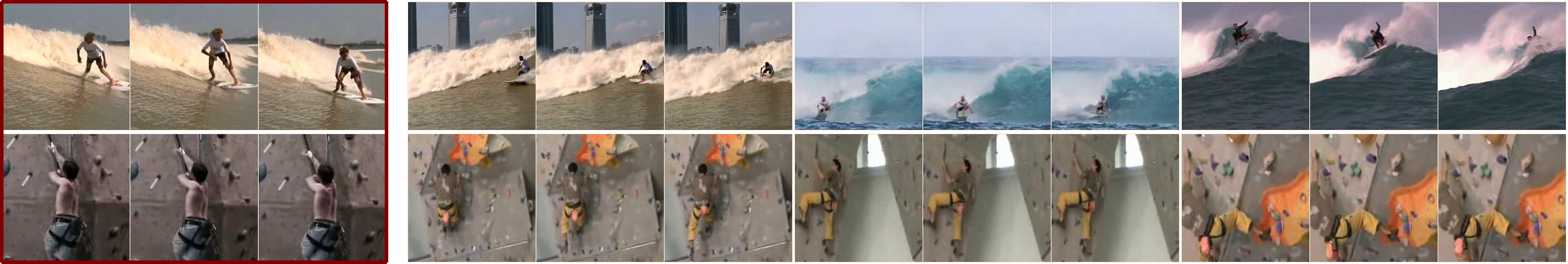} \\ 
			(a)\\
			\includegraphics[width=1.0\textwidth]{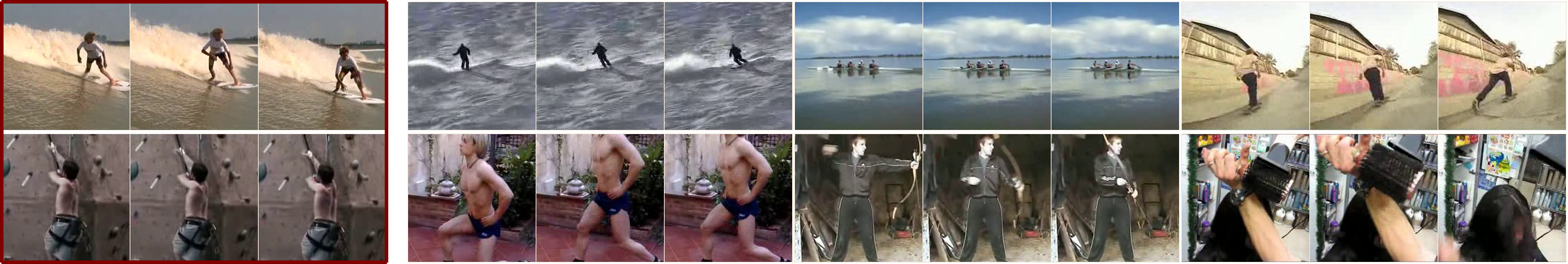}  \\
			\vspace{-0.1cm}(b)
		\end{tabular}
		\vspace{-2.5mm}
		\captionof{figure}{ Nearest Neighbour (NN) video clip retrieval on UCF101.  
			Each row contains four video clips, a query clip and the top three retrievals using clip embeddings.  
			To get the embedding,
			each video is passed to a 3D-ResNet18, average pooled to a single vector, 
			and cosine similarity is used for retrieval.
			(a) Embeddings obtained by Dense Predictive Coding (DPC); 
			(b) Embeddings obtained by using the inflated ImageNet pretrained weights. 
			The DPC captures the semantics of the human action, 
			rather than the scene appearance or layout as captured by the ImageNet trained embeddings.
			In the DPC retrievals the actual appearances of frames can vary dramatically, 
			\eg in the change in camera viewpoint for the climbing case.}
		\label{fig:teaser}
	\end{center}%
}]


\providecommand{\eg}[0]{e.g\xperiod}
\providecommand{\ie}[1]{i.e\xperiod}
\providecommand{\etal}[2]{et al.\xperiod}

\begin{abstract}
	\vspace{-4mm}
	The objective of this paper is self-supervised learning of spatio-temporal embeddings from video,
	suitable for human action recognition.
	
	We make three contributions: First, we introduce the Dense
	Predictive Coding~(DPC) framework for self-supervised representation
	learning on videos. 
	This learns a dense encoding of spatio-temporal blocks by recurrently predicting future representations;
	Second, we propose a curriculum training scheme to predict further into the future with progressively less temporal context.
	This encourages the model to only encode slowly varying spatial-temporal signals, therefore leading to semantic representations;
	Third, we evaluate the approach by first training the DPC model on the Kinetics-400 dataset with self-supervised learning, 
	and then finetuning the representation on a downstream task, \ie action recognition.
	With single stream~(RGB only), 
	DPC pretrained representations achieve state-of-the-art self-supervised performance on both UCF101~($75.7\%$ top1 acc) and HMDB51~($35.7\%$ top1 acc), outperforming all previous learning methods by a significant margin, and
	approaching the performance of a baseline pre-trained on ImageNet.
	The code is available at~\small{\url{https://github.com/TengdaHan/DPC}}.
	\vspace{-2mm}
\end{abstract}

\vspace{-3mm}
\section{Introduction} 
Videos are very appealing as a data source for self-supervision: there is almost an infinite supply available
(from Youtube etc.); image level proxy losses can be used at the frame level; and, there are plenty of additional
proxy losses that can be employed from the temporal information. 
One of the most natural, and consequently one of the first video proxy losses, is to predict future frames in the videos based on frames in the past. 
This has ample scope for exploration by varying the extent of the past knowledge (the temporal aggregation window 
used for the prediction) and also the temporal distance into the future for the predicted frames. 
However, future frame prediction does have a serious disadvantage -- that the future is not deterministic --  so methods
may have to consider multiple hypotheses with multiple instance losses, 
or other distributions and losses over their predictions.

Previous approaches to future frame prediction in video~\cite{Lotter17,Mathieu16,Srivastava15,Vondrick16a,Vondrick16b}
can roughly be
divided into two types: those that predict a reconstruction of the  
actual frames~\cite{Lotter17,Mathieu16,Srivastava15,Vondrick16a};
and those that only predict the latent representation (the embedding) of the frames~\cite{Vondrick16b}.
If our goal of self-supervision is only to learn a representation
that allows generalization for downstream discriminative tasks, \eg action recognition in video, 
then it may not be necessary to waste model capacity on resolving the stochasticity of frame appearance in detail, 
\eg appearance changes due to shadows, illumination changes, camera motion, etc. 
Approaches that only predict the frame embedding, 
such as Vondrick \etal~\cite{Vondrick16b}, avoid this potentially unnecessary task of detailed reconstruction, 
and use a mixture model to resolve the uncertainty in future prediction.
Although not applied to videos (but rather to speech signals and images), 
the Contrastive Predictive Coding (CPC) model of Oord \etal~\cite{Oord18} also learns embeddings, 
in their case by using a multi-way classification over temporal audio frames (or image patches), 
rather than the regression loss of~\cite{Vondrick16b}. 

In this paper we propose a new idea for learning spatio-temporal video embeddings, that we term ``Dense Predictive Coding'' (DPC). 
The model is designed to predict the future representations based on the recent past~\cite{Wiskott02}.
It is inspired by the CPC~\cite{Oord18} framework, 
and more generally by previous research on learning word embeddings~\cite{Mikolov13a,Mikolov13b,Mnih13a}.
DPC is also trained by using a variant of noise contrastive estimation~\cite{Gutmann10},
therefore, in practice, the model has never been optimized to predict the exact future, 
it is only asked to solve a multiple choice question, \ie pick the correct future states from lots of distractors. 
In order to succeed in this task, 
the model only needs to learn the shared semantics of the multiple possible future states, 
and this common/shared representation is the kind of invariance required in many of the vision tasks, \eg action recognition in videos. 
In other words,
the optimization objective will actually benefit from the fact that the future is not deterministic,
and map the representation of all possible future states to a space that their embeddings are close.
Concurrent work~\cite{Anand19} applies similar method on reinforcement learning. 

The contributions of this paper are three-fold:
First, we introduce Dense Predictive Coding~(DPC) framework for self-supervised representation learning on videos,
we task the model to predict the future embedding of the spatio-temporal blocks recurrently~(as used in N-gram prediction).
The model is trained to pick the ``correct'' future states from a pool of distractors, 
therefore treated as a multi-way classification problem.
Second, we propose a curriculum training scheme that enables the model to gradually predict further in the future 
(up to 2 seconds)
with progressively less temporal context,
leading more challenging training samples, 
and preventing the model from using shortcuts such as optical flow; 
Third, we evaluate the approach by first training the DPC model on the Kinetics-400~\cite{Kay17} dataset using self-supervised learning,
and then fine-tuning on action recognition benchmarks. 
Our DPC model achieves state-of-the-art self-supervised performance on both UCF101~($75.7\%$ top1 acc) and HMDB51~($35.7\%$ top1 acc), 
outperforming all previous single-stream~(RGB only) self-supervised learning methods by a significant margin.

\vspace{-3mm}

\section{Related Work}
\label{sec:relatedworks}
\noindent \textbf{Self-supervised learning from images.}
In recent years, methods for self-supervised learning on images have achieved an 
impressive performance in learning high-level image representations.
Inspired by the variants of Word2vec~\cite{Bengio03, Mikolov13a, Mikolov13b} that rely on predicting words
from their context, Doersch \etal~\cite{Doersch15} proposed the
pretext task of predicting the relative location of image patches.
This work spawned a line of work in context-based self-supervised
visual representation learning methods, \eg in ~\cite{Noroozi16}.
In contrast to the context-based idea, another set of pretext tasks
include carefully designed image-level classification, such as rotation~\cite{Gidaris18}
or pseudo-labels from clustering~\cite{Caron18}.
Another class of pre-text tasks is for 
dense predictions,
\eg image inpainting~\cite{Pathak16}, image colorization~\cite{Zhang16color}, and motion segmentation prediction~\cite{Pathak17}. 
Other methods instead enforce structural constraints on the representation space~\cite{Noroozi17}.
\\[5pt]
\noindent \textbf{Self-supervised learning from videos.}
Other than the predictive tasks reviewed in the introduction, another class of proxy tasks is based
on temporal sequence ordering of the frames~\cite{Misra16,Fernando17,Wei18}.
\cite{Isola15,Jayaraman16,Wang15} use the temporal coherence as a proxy loss.
Other approaches use egomotion~\cite{Agrawal15,Jayaraman15} to enforce equivariance in feature space \cite{Jayaraman15}. 
In contrast, \cite{Jing18} predicts the transformation applied to a spatio-temporal block. 
In~\cite{Kim19}, the authors propose to use a 3D puzzle as the proxy loss.
Recently~\cite{Vondrick18, Lai19, Wang19d}, leveraged  the natural temporal coherency of color in videos, to train
a network for tracking and correspondence related tasks.
\\[5pt]
\noindent \textbf{Action recognition with two-stream architectures.}
Recently, the two-stream architecture~\cite{Simonyan14b} has been a foundation for many competitive methods.
The authors show that optical flow is a powerful representation that improves action recognition dramatically.
Other modalities like audio signal can also benefits visual representation learning~\cite{Korbar18}.
While in this paper, 
we deliberately avoid using any information from optical flow or audio, 
and aim to probe the upperbound of self-supervised learning with \emph{only} RGB streams.
We leave it as a future work to explore how much boost optical flow branch and audio branch can bring to our self-supervised learning architecture.


\begin{figure*}[hbt!]
	\includegraphics[width=1.0\textwidth]{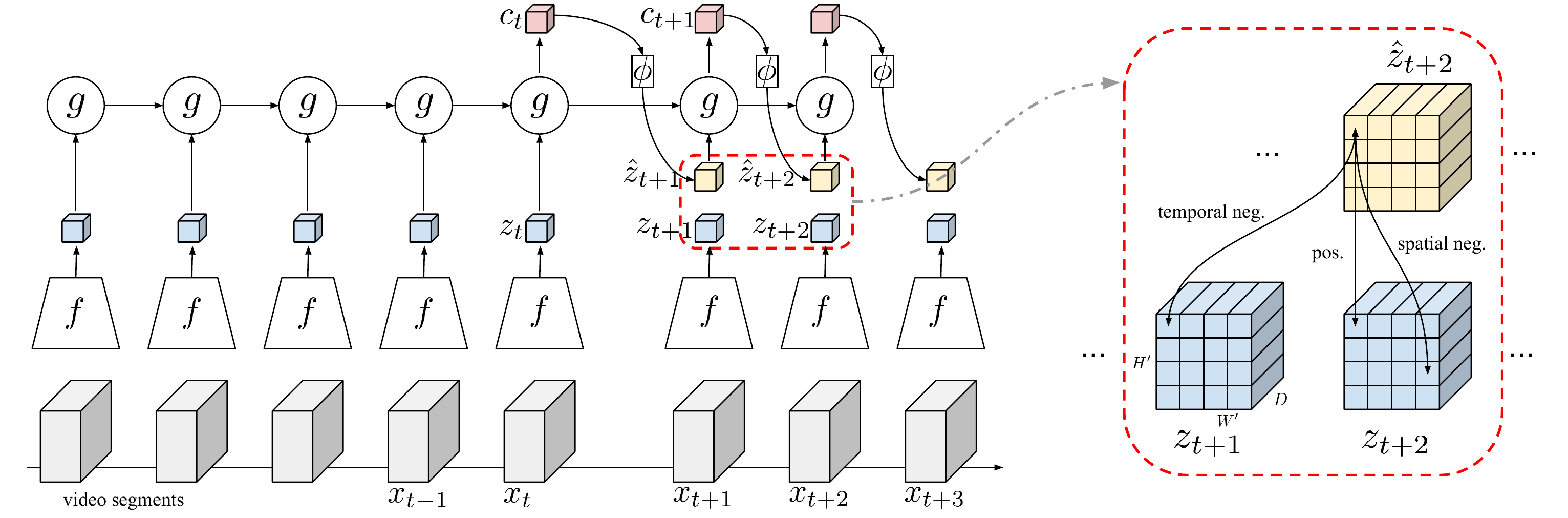}
	\centering
	\caption{A diagram of \textbf{Dense Predictive Coding} method. 
		The left part is the pipeline of the DPC, which is explained in Sec.~\ref{sec:method-framework}. 
		The right part (in the dashed rectangle) is an illustration of the Pred-GT pair construction for contrastive loss, 
		which is explained in Sec.~\ref{sec:method-loss}.}
	\vspace{-4mm}
	\label{fig:arch}
\end{figure*}

\vspace{-2mm}
\section{Dense Predictive Coding~(DPC)}
\label{sec:method}
In this section, we describe the learning framework, 
details of the architecture,  and the curriculum training that gradually learns to predict further into  the future with progressively less temporal context.

\subsection{Learning  Framework}
\label{sec:method-framework}
The goal of DPC is to predict a  slowly varying semantic representation based on the recent past, 
\eg we construct a prediction task that observes about 2.5 seconds of the video and predict the embedding for the future 1.5 seconds,
as illustrated in Figure~\ref{fig:arch}. 
A video clip is partitioned into multiple non-overlapping blocks \(x_{1}, x_{2}, \dots, x_{n}\),
with each block containing an equal number of frames.
First, a non-linear encoder function \(f(.)\) maps each input video block \(x_{t}\) to its latent representation $z_t$, 
then an aggregation function \(g(.)\) temporally aggregates \(t\) consecutive latent representations into a context representation $c_t$:
\begin{align}
z_t = f(x_t) \\
c_t = g(z_1, z_2, ..., z_t)
\end{align}
where $x_t$ has dimension $\mathbb{R}^ {T \times H \times W \times C}$, 
and $z_t$ is a feature map with dimension $\mathbb{R}^ {1 \times H' \times W' \times D}$, 
organized as time $\times$ height  $\times$ width $\times$ channels.
\footnote{In our initial experiments,  
	$x_t \in \mathbb{R}^{5 \times 128 \times 128 \times 3}$, 
	$z_t \in \mathbb{R}^{1 \times 4 \times 4 \times 256}$}

The intuition behind the predictive task is that if one can infer future semantics from \(c_t\), 
then the context representation \(c_t\) and the latent representations \(z_1, z_2, ..., z_t\) must have encoded strong semantics of the input video clip. Thus, 
we introduce a predictive function \(\phi(.)\) to predict the future. 
In detail, \(\phi(.)\) takes the context representation as the input and predicts the future clip representation:
\begin{align}
\hat{z}_{t+1} = \phi(c_{t}) = \phi\big(g(z_1, z_2, \dots, z_t)\big) \\
\hat{z}_{t+2} = \phi(c_{t+1}) = \phi\big(g(z_1, z_2, \dots, z_t, \hat{z}_{t+1})\big)\label{eq:seq}
\end{align}
where \(c_{t}\) denotes the context representation from time step \(1\) to \(t\), 
and \(\hat{z}_{t+1}\) denotes the predicted latent representation of the time step \(t+1\). 
In the spirit of Seq2seq~\cite{Sutskever14}, 
representations are predicted in a sequential manner.
We predict \(q\) steps in the future,
at each time step \(t\), the model consumes the previously generated embedding~($\hat{z}_{t-1}$) as input 
when generating the next~($\hat{z}_t$),
further enforcing  the prediction to be conditioned on all previous observations and predictions, 
and therefore encourages  an N-gram like video representation.

\subsection{Contrastive Loss}
\vspace{-1mm}
\label{sec:method-loss}
Noise Contrastive Estimation (NCE)~\cite{Gutmann10} constructs a binary classification task: 
a classifier is fed with real samples and noise samples, 
and the objective is to distinguish them. A variant of NCE~\cite{Mnih13a, Oord18} classifies one real sample among many noise samples. Similar to~\cite{Mnih13a, Oord18}, we use a loss based on NCE for the predictive task. 
NCE over feature embeddings encourages the predicted representation \(\hat{z}\) to be close to the ground truth representation \(z\), 
but not so strictly that it has to resolve the low-level stochasticity.

In the forward pass, the ground truth representation \(z\) and the predicted representation \(\hat{z}\) are computed. 
The representation for the $i$-th time step is denoted as \(z_{i}\) and \(\hat{z}_{i}\), which have the same dimensions.
Note that, instead of pooling into a feature vector, 
both \(z_{i}\) and \(\hat{z}_{i}\) are kept as feature maps~(\(z_{i},\hat{z}_{i}\in \reals{H' \times W' \times D}\)),
which maintains the spatial layout representation. 
We denote the feature vector in each spatial location of the feature map as \(z_{i, k}\in \reals{D}\) and \(\hat{z}_{i, k}\in \reals{D}\) 
where \(i\) denotes the temporal index and \(k\) is the spatial index \(k\in \set{(1,1), (1,2),\dots, (H, W)}\). 
The similarity of the predicted and ground-truth pair (Pred-GT pair) is computed by the dot product \(\transpose{\hat{z}_{i,k}}{z}_{j,m}\). The objective is to optimize:
\vspace{-1mm}
\begin{equation}
\mathcal{L}=-\sum_{i,k}\Bigg[\log{\frac{ \exp(\transpose{\hat{z}}_{i,k} \cdot {z}_{i,k} ) }
	{\sum_{j,m}{\exp(\transpose{\hat{z}}_{i,k} \cdot {z}_{j,m})}}}\Bigg] 
\end{equation}\label{eq:loss}

In essense, this is simply a cross-entropy loss~(negative log-likelihood) 
that distinguishes the positive Pred-GT pair out of all other negative pairs. 
For a predicted feature vector \(\hat{z}_{i,k}\), 
the only positive pair is \((\hat{z}_{i,k}, {z}_{i,k})\), i.e.\ the predicted and ground-truth features at the same time step and same spatial location. 
All the other pairs \((\hat{z}_{i,k}, {z}_{j,m})\) where \((i,k)\neq (j,m)\), are negative pairs. 
The loss encourages the positive pair to have a higher similarity than any negative pairs. 
If the network is trained in a mini-batch consisting of \(B\) video clips and each of the \(B\) clips is from distinct video, 
more negative pairs can be obtained. 

To discriminate the different types of negative pairs, given a Pred-GT pair \((\hat{z}_{i,k}, {z}_{j,m})\), 
we define the terminology as follows: 

\vspace{-3.5mm}
\paragraph{Easy negatives:} is the Pred-GT pair that is formed from two distinct videos. 
These pairs are naturally easy because they usually have distinct color distributions and thus predicted feature and ground-truth feature have low similarity. 

\vspace{-3.5mm}
\paragraph{Spatial negatives:} 
is the Pred-GT pair that is formed from the same video but at a different spatial position in the feature map, 
\ie \(k\neq m\), while $i,j$ can be any index.

\vspace{-3.5mm}
\paragraph{Temporal negatives (hard negatives):} is the Pred-GT pair that comes from the same video and same spatial position, 
but from different time steps, \ie \(k=m, i\neq j\). 
They are the hardest pair to classify because their score will be very close to the positive pairs.

\vspace{1mm}
Overall, we use a similar idea to the Multi-batch training~\cite{Tadmor16}. If the mini-batch has batch size $B$, 
the feature map has spatial dimension $ H' \times W'$ and the task is to classify one of $ q $ time steps,
the number of each classes follows:
\vspace{-1mm}
\begin{equation*}
\begin{aligned}
& \text{Pos} : \text{N}_{temporal} : \text{N}_{spatial} : \text{N}_{easy} \\
= & 1 : (q-1) : (H'W' -1)q : (B-1)H'W' q
\end{aligned}
\end{equation*}


\vspace{-3.5mm}
\paragraph{Curriculum learning strategy.} 
A curriculum learning strategy is designed by progressively increasing
the number of prediction steps of the model (Sec.~\ref{exp:step}).
For instance, the training process can start by  predicting only 2
steps (about 1 second), \ie only computing \(\hat{z}_{t+1}\) and \(\hat{z}_{t+2}\), and
the Pred-GT pairs are constructed between \(\{ z_{t+1},z_{t+2} \}\)
and \(\{ \hat{z}_{t+1}, \hat{z}_{t+2} \}\).  After the network has
learnt this simple task, it can be trained to predict 3 steps (about 1.5 seconds), \eg
computing \(\hat{z}_{t+1}\), \(\hat{z}_{t+2}\) and \(\hat{z}_{t+3}\)
and construct Pred-GT pairs accordingly.  Importantly, curriculum
learning introduces more hard negatives throughout the training
process, and forces  the model to gradually learn  to predict further in
the future with progressively less temporal context.  Meanwhile, the
model is gradually trained to grasp the uncertain nature in its
prediction.

\subsection{Avoiding Shortcuts and Learning Semantics}
\label{sec:method-sc}

Empirical experience in self-supervised learning indicates that if the
proxy task is well-designed and requires semantic understanding, a
more difficult learning task usually leads to a better-quality
representation~\cite{Lee17}.  However, ConvNets are notoriously known
for learning shortcuts for tackling tasks~\cite{Doersch15,Noroozi16,Wei18}.
In our training, we employ a number of mechanisms to avoid potential
shortcuts, as detailed next.

\vspace{-4.5mm}
\paragraph{Disrupting optical flow.}
A trivial solution of our predictive task is that \(f(.)\), \(g(.)\)
and $\phi(.)$ together learn to capture low-level optical flow
information and perform feature extrapolation as the prediction.  
To force the model to learn high-level semantics, 
a critical operation is frame-wise augmentation,
\ie random augmentation for each individual frame in the video blocks, 
such as frame-wise color jittering including random brightness, contrast,
saturation, hue and random greyscale during  training.
Furthermore,  the curriculum of predicting  further into the future,
\ie predicting  the semantics for the next a few seconds, 
also ensures that  optical flow alone will not be able to solve this  prediction task. 

\vspace{-4.5mm}
\paragraph{Temporal receptive field.}
The temporal receptive field (RF) of \(f(.)\) is limited by cutting the 
input video clip into non-overlapping blocks before feeding it into \(f(.)\). 
Thus, the effective temporal RF of each feature map \(z_{i}\) is strictly restricted to be within each  video block. 
This  avoids the network being able 
to discriminate positive and hard-negative by recognizing relative temporal position.

\vspace{-4.5mm}
\paragraph{Spatial receptive field.}
Due to the depth of CNN, 
each feature vector \(\hat{z}_{i,k}\) in the final predicted feature map \(\hat{z}_i\) has a large spatial RF that
(almost) covers the entire input spatial dimension. 
This creates a shortcut to discriminate positive and spatial negative by using padding patterns. 
One can limit the spatial RF by cutting input frames into patches~\cite{Oord18,Kim19}.  
However this brings some drawbacks:
First, the self-supervised pre-trained network will have limited receptive field~(RF), 
so the representation may not generalize well for downstream tasks where a large RF is required.  
Second, limiting spatial RF in videos makes the context feature too weak.  
The context feature has a spatio-temporal RF that covers a thin cube in the video flow.  
Neglecting context  is also not ideal for understanding video semantics and brings ambiguity to the predictive task.  
Considering this trade-off, our method does not restrict the spatial RF.

\vspace{-4.5mm}
\paragraph{Batch normalization.}
Common practice uses Batch Normalization~\cite{Ioffe15} (BN) in deep CNN architecture. 
The BN layer may provide shortcuts that the network acknowledges the statistical distribution of the mini-batch,
which benefits the classification. 
In~\cite{Oord18}, the authors demonstrate BN results in network cheating, 
and the ResNet trained with BN does not generalize to the downstream image classification task.
In our method, we find the effect of BN shortcut is very limited. 
The self-supervised training gives similar accuracy using either BN or Instance Normalization~\cite{Ulyanov16} (IN). 
For downstream tasks like classification, a network with BN gives 5\%-10\% accuracy gain comparing with a network with IN. 
It is hard to train a deep CNN without normalization for either self-supervised training or supervised training. 
Overall, we use BN in our encoder function \(f(.)\). 


\vspace{5pt}
\subsection{Network Architecture}
We choose to use a 3D-ResNet similar to~\cite{Hara18} as the encoder \(f(.)\).  
Following the convention of~\cite{Feichtenhofer18} there are four residual blocks in ResNet architecture, 
namely \(\text{res}_2\), \(\text{res}_3\), \(\text{res}_4\) and \(\text{res}_5\),
and only expand the convolutional kernels in \(\text{res}_4\) and \(\text{res}_5\) to be 3D ones.  
For experiment analysis, we used 3D-ResNet18, denoted as R-18 below.

To train a strong encoder \(f(.)\), a weak aggregation function \(g(.)\) is preferable. 
Specifically, a one-layer Convolutional Gated Recurrent Unit (ConvGRU) with kernel size \((1,1)\) is used, 
which shares the weights amongst all spatial positions in the feature map. 
This design allows the aggregation function to propagate features in the temporal axis. 
A dropout~\cite{Srivastava14} with \(p=0.1\) is used when computing hidden state in each time step. 
A shallow two-layer perceptron is used as the predictive function \(\phi(.)\). 

\vspace{-1mm}
\subsection{Self-Supervised Training}
For data pre-processing, 
we use 30 fps videos with a uniform temporal downsampling by factor 3, \ie take one frame from every 3 frames. 
These consecutive frames are grouped into 8 video blocks where each block consists of 5 frames. 
Frames are sampled in a consecutive way with consistent temporal stride to preserve the temporal regularity, 
because random temporal stride introduces uncertainties to the predictive task especially 
when the network needs to distinguish the difference among different time steps. 
Specifically, each video block spans over 0.5s and the entire 8 segments span over 4s in the raw video. 
The predictive task is initially designed to observe the first 5 blocks and predict the remaining 3 blocks (denoted as `5pred3' afterwards), which is observing 2.5 seconds to predict the following 1.5 seconds. 
We also experiment with different predictive configuration like 4pred4 in Sec.~\ref{exp:step}.

For data augmentation, 
we apply random crop, random horizontal flip, random grey, and color jittering. 
Note that the random crop and random horizontal flip are applied for the entire clip in a consistent way.
Random grey and color jittering are applied in a frame-wise manner to prevent the network from learning low-level flow information as mentioned above~(in Sec.~\ref{sec:method-sc}), 
\eg each video block may contain both colored and grey-scale image with different contrast.
All models are trained end-to-end using Adam~\cite{KingmaB14} optimizer 
with an initial learning rate \(10^{-3}\) and weight decay \(10^{-5}\). 
Learning rate is decayed to \(10^{-4}\) when validation loss plateaus. 
A batchsize of 64 samples per GPU is used, and our experiments use 4 GPUs. 


\vspace{-3mm}
\section{Experiments and Analysis}
\label{sec:exp}

In the following sections we present controlled experiments, 
and aim to investigate four aspects:
\emph{First}, an ablation study on the DPC model to show the function of different design choices, 
\eg sequential prediction, dense prediction.
\emph{Second}, the benefits of training on a larger, and more diverse dataset.
\emph{Third}, the correlation between performance on 
self-supervised learning and performance on the downstream supervised learning task.
\emph{Fourth}, the variation in the learnt representations when predicting further into the future.

\vspace{-4mm}
\paragraph{Datasets.}
The DPC is a general self-supervised learning framework for any video types, 
but we focus here on human action videos
\eg UCF101~\cite{Soomro12}, HMDB51~\cite{Kuehne11} and Kinetics-400~\cite{Kay17} datasets.
UCF101 contains 13K videos spanning over 101 human action classes.
HMDB51 contains 7K videos from 51 human action classes.  
Kinetics-400 (K400) is a big video dataset containing 306K video clips for 400 human
action classes.  

\vspace{-4mm}
\paragraph{Evaluation methodology.}
The self-supervised model is trained either on UCF101 or K400.
The representation is evaluated by its performance 
on a downstream task, \ie action
classification on UCF101 and HMDB51.
For all the experiments below:
we report top1 accuracy for self-supervised  learning in the middle column of all tables;
and report the top1 accuracy for supervised learning for action classification on UCF101
in the rightmost column.
In self-supervised learning, 
the top1 accuracy refers to how often the multi-way classifier picks the right Pred-GT pair, 
\ie this is not related with any action classes.
While for supervised learning, the top1 accuracy indicates the action classification accuracy on UCF101.
Note,  we
report the first training/testing splits of UCF101 and HMDB51 in
all the experiments, apart from the  comparison with the state of the art 
in Table~\ref{table:sota} where we report the average accuracy over three splits.

\vspace{-4mm}
\paragraph{Action classifier.}
During supervised learning, 
$5$ video blocks are passed as input~(the same as for self-supervised training, 
\ie each block is of $\mathbb{R}^{5 \times 128 \times 128 \times 3}$),
and encoded as a sequence of feature maps with the encoding function $f(.)$ (a 3D-ResNet). 
As with the self-supervised architecture, the aggregation function $g(.)$ (a ConvGRU) aggregates the feature maps over time and produces a context feature. 
The context feature is further passed through a spatial pooling layer followed by a fully-connected layer 
and a multi-way softmax for action classification.
The classifier is trained using the Adam~\cite{KingmaB14} optimizer with an initial learning rate \(10^{-3}\) and weight decay \(10^{-3}\). 
Learning rate is decayed twice to \(10^{-4}\) and \(10^{-5}\)
Note that the \emph{entire} network is trained end-to-end. 
The details of the architecture are given in Appendix~\ref{sec:arch}.

During inference, 
video clips from the validation set are densely sampled from an input video 
and cut into blocks~($\mathbb{R}^{5 \times 128 \times 128 \times 3}$) with half-length overlapping.
Augmentations are removed and only center crop is used. 
The softmax probabilities are averaged to give the final classification result. 

\subsection{Performance Analysis}
\subsubsection{Ablation Study on Architecture}
\label{sec:exp-vs_cpc}
In this section, we present an ablation study by gradually removing components from the DPC model (see Table~\ref{table:vs-cpc}).
For efficiency, all the self-supervised learning experiments refer to the 5pred3 setting, 
\ie 5 video blocks~(2.5 second) are used as input to predict the future 3 steps~(1.5 second).

\begin{table}[htbp]
	\centering
	\small
	\begin{tabular}{lllc|c}
		\hline
		\multicolumn{1}{c}{\multirow{2}{*}{Network}}   & \multicolumn{3}{c}{Self-Sup. (UCF)}    & \multicolumn{1}{|c}{Sup. (UCF)}  \\
		\multicolumn{1}{c}{}       & setting  & method            & top1 acc         & top1 acc          \\  \hline
		R-18 &  -     & - (rand. init.)       & -                       &  46.5      \\ \hline
		R-18 &  5pred3     & DPC                     & \textbf{53.6}   &  \textbf{60.6}\\
		R-18 &  5pred3     & remove Seq.       & 51.3                 &  56.9         \\
		R-18 &  5pred3     & remove Map        & 36.5                &  44.9         \\
		\hline
	\end{tabular}
	\caption{Ablation study of DPC. 
		\emph{remove Seq} means removing the sequential prediction mechanism in DPC, and replacing by parallel prediction.
		\emph{remove Map} means removing the dense feature map design in DPC, and use a feature vector instead.
		Self-supervised tasks are trained on UCF101 using 5pred3 setting. Representation learned from each self-supervised task is evaluated by training a supervised action classifier on UCF101. 
	}\label{table:vs-cpc}
\end{table}

Compared with the baseline model trained with random initialization and fully supervised learning, 
our DPC model pre-trained with self-supervised learning has a significant boost~(top1 acc: 46.5\% vs. 60.6\%).
When removing the sequential prediction, 
\ie all 3 future steps are predicted in parallel with three different fully-connected layers, 
the accuracy for both self-supervised learning and supervised learning start to drop.
Lastly, we further replace the dense feature map by the average-pooled feature vector, \ie it becomes a CPC-like model,
we are not able to train this model either on self-supervised learning task or supervised learning.
This demonstrates that \emph{dense} predictive coding is essential to our success, 
and sequential prediction also helps to boost the model performance.

\subsubsection{Benefits of Large Datasets}
\label{exp:scale}

In this section, 
we investigate the benefits of pre-training on a large-scale dataset~(UCF101~vs.~K400),
we keep the 5pred3 setting and evaluate the effectiveness for downstream task on UCF101.
Results are shown in Table~\ref{table:big-data}. 

\begin{table}[htbp]
	\centering
	\small
	\begin{tabular}{lllc|c}
		\hline
		\multicolumn{1}{c}{\multirow{2}{*}{Network}}& \multicolumn{3}{c}{Self-Sup.} & \multicolumn{1}{|c}{Sup. (UCF)} \\
		\multicolumn{1}{c}{}           & setting                & dataset           & top1 acc     & top1 acc                   \\ \hline
		R-18                      & 5pred3               & UCF101         & 53.6         & 60.6                     \\
		R-18                      & 5pred3               & K400           & \textbf{61.1}& \textbf{65.9}            \\ \hline
	\end{tabular}
	\caption{Results of DPC on UCF101 and K400 respectively. Both experiments use 5pred3 setting. Representations are evaluated by training a supervised action classifier on UCF101 (right column).}\label{table:big-data}
\end{table}

Training the model on K400 increases the self-supervised accuracy to 61.1\%,
and supervised accuracy from $60.6\%$ to $65.9\%$,
suggesting the model has captured more regularities than a smaller dataset like UCF101. 
It is clear that DPC will benefit from large-scale video dataset~(infinite supply available), 
which naturally provides more diverse negative Pred-GT pairs.

\subsubsection{Self-Supervised vs. Classification Accuracy}
\label{sec:exp-acc_vs_acc}
\begin{figure}[htbp]
	\centering
	\includegraphics[width=0.45\textwidth]{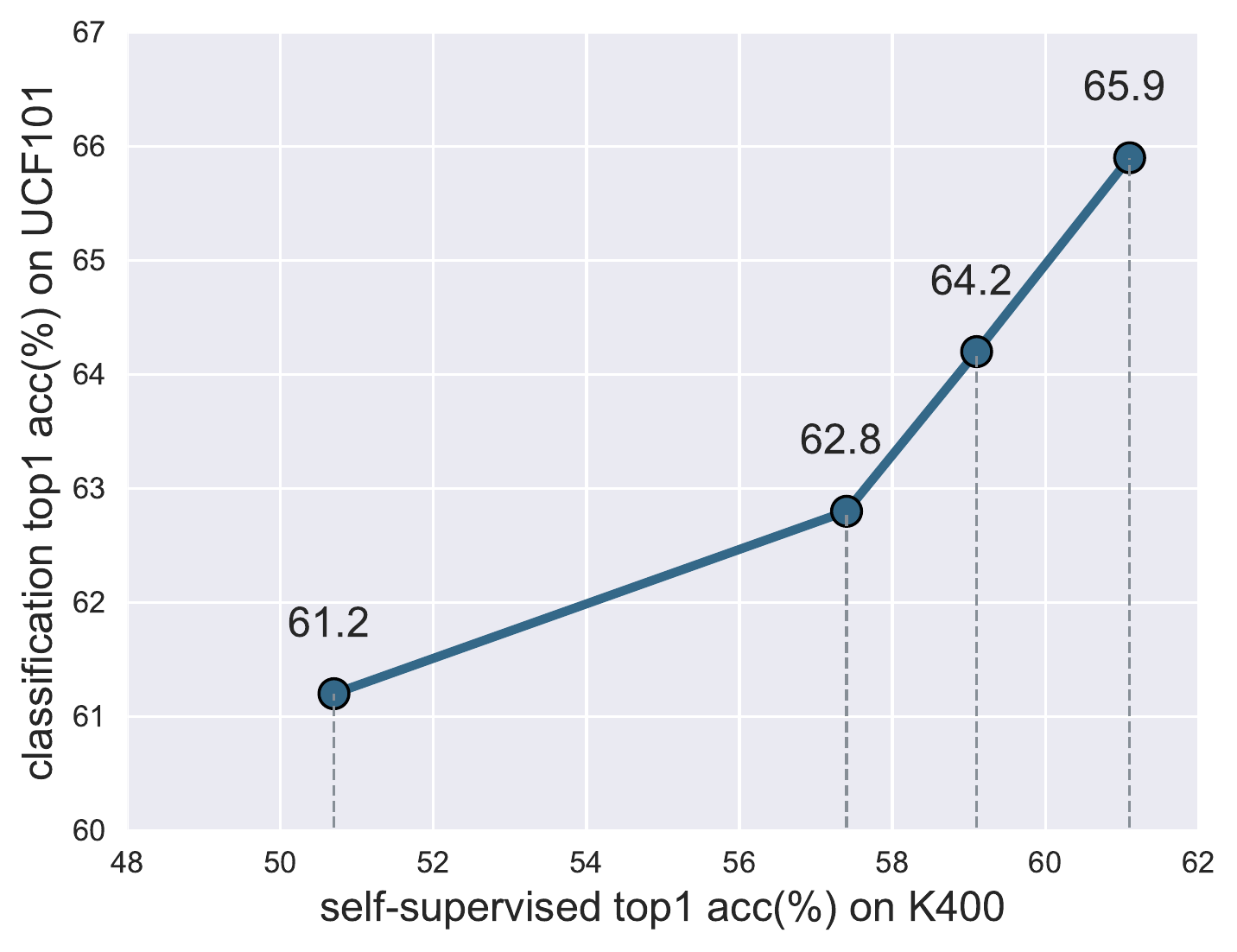}
	\vspace{-3mm}
	\caption{Relation between self-supervised accuracy and classification accuracy. 
		Self-supervised model (DPC) is trained on K400 and the weights at epoch $\{13,48,81,109\}$ are saved, 
		which achieve $\{50.7\%, 57.4\%, 59.1\%, 61.1\%\}$ self-supervised accuracy respectively. 
		The checkpoints are evaluated by finetuning on UCF101. }
	\label{fig:acc-vs-acc}
\end{figure}

In this section, 
we investigate the correlation between the accuracy of self-supervised learning and downstream supervised learning.
While training DPC~(5pred3 task on K400),  
we evaluate the representation at different training stages (number of epochs) on the downstream task~(on UCF101). 
The results are shown in Figure~\ref{fig:acc-vs-acc}.

It can be seen that a higher accuracy in self-supervised task always
leads to a higher accuracy in downstream classification.  The result
indicates that DPC has actually learnt visual representations
that are not only specific to self-supervised task, but are also generic
enough to be beneficial for the downstream task.

\subsubsection{Benefits of Predicting Further into the Future}
\label{exp:step}
Due to  the increase of uncertainty, predicting further into the  future  in video sequences gets more difficult, 
therefore more abstract~(semantic) understanding is required.
We hypothesize that if we can train the model to predict further, 
the learnt representation should be even better.
In this section, 
we employ curriculum learning to gradually train the model to predict further with progressively less temporal context,
\ie from 5pred3 to 4pred4~(4 video blocks as input and predict the future 4 steps).

\begin{table}[htbp]
	\centering
	\small
	\begin{tabular}{llcc|c}
		\hline
		\multicolumn{1}{l}{\multirow{2}{*}{Network}} & \multicolumn{3}{c}{Self-Sup. (K400)} & \multicolumn{1}{|c}{Sup. (UCF)}      \\
		\multicolumn{1}{l}{} & \multicolumn{1}{l}{setting} & \multicolumn{1}{l}{curr.} & \multicolumn{1}{l}{top1 acc} & \multicolumn{1}{|l}{top1 acc} \\
		\hline
		R-18     & 5pred3         & \xmark & 61.1          & 65.9      \\
		R-18     & 4pred4         & \xmark & 48.3          & 64.9  \\ 
		R-18     & 5pred3+4pred4  & \cmark & \textbf{50.8} & \textbf{68.2}  \\ 
		\hline
	\end{tabular}
	\caption{Results of DPC with different prediction steps. 
		All models are trained on K400 with \emph{same} number of 320k iterations. 
		Note that for 5pred3 and 4pred4, the model is trained from scratch. 
		`5pred3+4pred4' denotes that curriculum learning strategy, \ie initialized with the pre-trained weights from 5pred3 task. 
		The representation is evaluated by training an action classifier on UCF101 (right column).}
	\label{table:long-short}
	\vspace{-2mm}
\end{table} 

The result shows that the 4pred4 setting gives a substantially lower
accuracy on the self-supervised learning than 5pred3. This is
actually not surprising, as 4pred4 naturally introduces 33\% more hard
negative pairs than predicting future 3 steps, making the
self-supervised learning more difficult~(explained in
Section~\ref{sec:method-loss}).

Interestingly, despite a lower accuracy on self-supervised learning task, 
when comparing with 5pred3, curriculum learning on 4pred4 provides $2.3\%$ performance boost on the downstream supervised task~(top1 acc: 68.2\% vs. 65.9\%). The experiment also shows that curriculum learning is effective as it achieves higher performance than training 4pred4 task from scratch~(top1 acc: 68.2\% vs. 64.9\%). Similar effect is also observed in~\cite{Korbar18}.

\vspace{-2mm}
\subsubsection{Summary}
Through the experiments above, 
we have demonstrated the keys to the success of DPC.
\emph{First}, it is critical to do dense predictive coding, \ie predicting both temporal and spatial representation in the future blocks,
and sequential prediction enables a further boost in the quality of the learnt representation.
\emph{Second}, a large-scale dataset helps to improve the self-supervised learning, 
as it naturally contains more world patterns and provides more diverse negative sample pairs.
\emph{Third}, the representation learnt from DPC is generic, 
as a higher accuracy in the self-supervised task also yield a higher accuracy in the downstream classification task.
\emph{Fourth}, predicting further into the future is also beneficial, 
as the model is forced to encode the high-level semantic representations, 
and ignore the low-level information.

\section{Comparison with State-of-the-art Methods}

\begin{table*}[!htb]
	\centering
	\small
	\begin{tabular}{lll|ll}
		\hline
		\multicolumn{3}{c}{Self-Supervised Method~(RGB stream only)}              & \multicolumn{2}{|c}{Supervised Accuracy (top1 acc)} \\
		Method                          & Architecture (\#param)   & Dataset     & UCF101    & HMDB51        \\ 
		\hline
		
		Random Initialization           & 3D-ResNet18 (14.2M)    & -               & 46.5           &  17.1            \\ 
		ImageNet Pretrained~\cite{Simonyan14b}        & VGG-M-2048 (25.4M)   & -               & 73.0           &  40.5           \\  \hline
		
		Shuffle~\&~Learn~\cite{Misra16}~($227 \times 227$) & CaffeNet (58.3M)    & UCF101/HMDB51   & 50.2           &  18.1            \\
		OPN~\cite{Lee17}~($80 \times 80$)  & VGG-M-2048 (8.6M)  & UCF101/HMDB51   & 59.8           &  23.8            \\
		OPN~\cite{Lee17}~($120 \times 120$)  & VGG-M-2048 (11.2M)  & UCF101/HMDB51& 55.4           &  -            \\
		OPN~\cite{Lee17}~($224 \times 224$)  & VGG-M-2048 (25.4M)  & UCF101/HMDB51& 51.9           &  -            \\
		\textbf{Ours~($128 \times 128$)}                     & 3D-ResNet18 (14.2M) & UCF101    & \textbf{60.6} &  -  \\ 
		\hline
		3D-RotNet~\cite{Jing18}~($112 \times 112$)         & 3D-ResNet18-full (33.6M) & Kinetics-400    & 62.9           &  33.7            \\
		3D-ST-Puzzle~\cite{Kim19}~($224 \times 224$)     & 3D-ResNet18-full (33.6M) & Kinetics-400    & 63.9 (65.8\textsuperscript{$\star$})&  33.7\textsuperscript{$\star$}            \\
		\textbf{Ours~($128 \times 128$)}                   & 3D-ResNet18 (14.2M) & Kinetics-400    & \textbf{68.2} &  \textbf{34.5}  \\ 
		\textbf{Ours~($224 \times 224$)}                   & 3D-ResNet34 (32.6M) & Kinetics-400    & \textbf{75.7} &  \textbf{35.7}  \\ 
		\hline
	\end{tabular}
	\caption{Comparison with other self-supervised methods, results are reported as an average over three training-testing splits. 
		Note that, previous works~\cite{Jing18,Kim19} use full-scale 3D-ResNet18, \ie all convolutions are 3D,
		and the input sizes for different models have been shown.
		\textsuperscript{$\star$}indicates the results from the multi-task self-supervised learning, \ie Rotation + 3D Puzzle.}
	\label{table:sota}
\end{table*}


The results are given in Table~\ref{table:sota}, four phenomena can be observed:
\emph{First},
when self-supervised training with only UCF101, 
our DPC~(60.6\%) outperforms all previous methods under similar settings. 
Note that OPN~\cite{Lee17} performs worse when input resolution increases, 
which indicates a simple self-supervised task like order prediction may not capture the rich semantics from videos.
\emph{Second},
when using Kinetics-400 for self-supervised pre-training,
our DPC~(68.2\%) outperforms all the previous methods by a large margin. 
Note that, in the work~\cite{Jing18, Kim19}, 
the authors use a full-scale 3D-ResNet18 architecture~(33.6M parameters), \ie all convolutions are 3D,
however our modified 3D-ResNet18 has fewer parameters~(only the last 2 blocks are 3D convolutions). 
The authors of~\cite{Kim19} obtain 65.8\% accuracy by combing the rotation classification~\cite{Jing18} 
with their Space-Time Cubic Puzzles method, essentially multi-task learning.
When only considering their Space-Time Cubic Puzzles method, they obtain 63.9\% top1 accuracy. 
On HMDB51, our method also outperforms the previous state of the art result by 0.8\% (34.5\% vs. 33.7\%). 
\emph{Third},
when applying on larger input resolution ($224\times224$) and using model with more capacity (3D-ResNet34), 
our DPC clearly dominate all self-supervised learning methods~(75.7\% on UCF101 and 35.7\% on HMDB51),
further demonstrating that DPC is able to take advantage from networks with more capacity and today's large-scale datasets.
\emph{Fourth},
ImageNet pretrained weights have been a golden baseline for action recognition~\cite{Simonyan14b},
our self-supervised DPC is the first model that surpasses the performance of models~(VGG-M) pre-trained with ImageNet~($75.7\%$ vs. $73.0\%$ on UCF101).

\vspace{3pt}
\subsection{Visualization}

\begin{figure*}[h]
	\centering
	\begin{subfigure}{1.0\textwidth}
		\centering
		\includegraphics[width=1.0\linewidth]{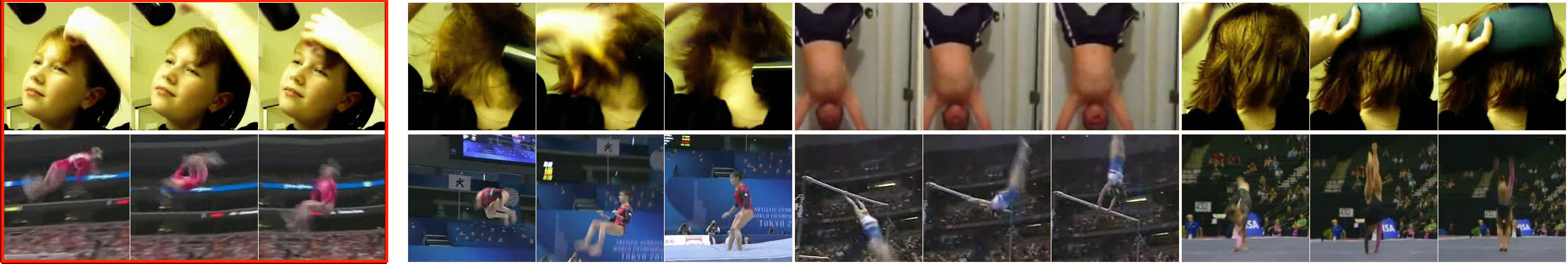}
		\caption{}
		\label{fig:vis1}
	\end{subfigure}
	
	\vspace{5pt}
	\begin{subfigure}{1.0\textwidth}
		\centering
		\includegraphics[width=1.0\linewidth]{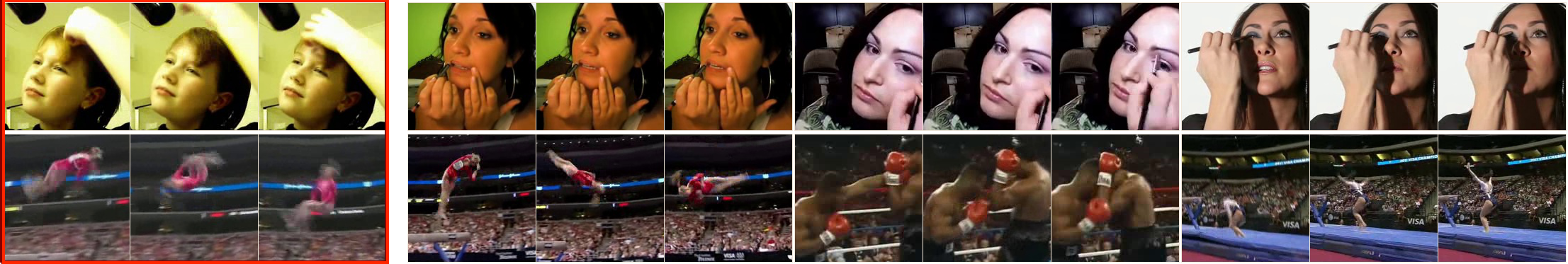}
		\caption{}
		\label{fig:vis2}
	\end{subfigure}
	\vspace{-5pt}
	\caption{More examples of video retrieval with nearest neighbour (same setting as Figure~\ref{fig:teaser}). 
		Figure~\ref{fig:vis1} is the NN retrieval with DPC pre-trained \(f(.)\) on UCF101 (performance reported in Sec.~\ref{exp:scale}). 
		Figure~\ref{fig:vis2} is the NN retrieval with ImageNet inflated \(f(.)\). Retrieval is performed on UCF101 validation set.}
	\label{fig:vis}
\end{figure*}

We visualize the Nearest Neighbour (NN) of the video segments in the
spatio-temporal feature space in Figure~\ref{fig:vis} and Figure~\ref{fig:teaser}.  
In detail, one video segment is randomly sampled from each video, 
then the spatio-temporal feature \(z_{i}=f(x_{i})\) is extracted and pooled into
a vector.  Then the feature vector is used to compute the cosine similarity score.  
In all figures, Figure~\ref{fig:vis1} includes the video clips retrieved using our DPC model from self-supervised learning, 
note that the network does not receive any class label information during training.  
In comparison, Figure~\ref{fig:vis2} uses the inflated ImageNet pre-trained weights.

It can be seen, that the ImageNet model is able to encode the scene semantics, \eg human faces, crowds, 
but does not capture any semantics about the human actions.
In contrast, our DPC model has actually learnt the video semantics \emph{without} using any manual annotation, 
for instance, despite the background change in running, DPC can still correctly retrieve the video block.

\vspace{3pt}
\subsection{Discussion}
Why should the DPC model succeed in learning a representation suitable for action recognition, 
given the problem of a non-deterministic future?  
There are three reasons: 
First, the use of the softmax function and multi-way classification loss enables multi-modal,
skewed, peaked or long tailed distributions; 
the model can therefore handle the task of predicting the non-deterministic future.
Second, by avoiding the shortcuts, the model has been prevented from learning
simple smooth extrapolation of the embeddings; 
it is forced to learn semantic embeddings to succeed in its learning task.
Third, in essense, DPC is trained by predicting future representations, 
and use them as a ``query'' to pick the correct ``key'' from lots of distractors. 
In order to succeed in this task, 
the model has to learn the shared semantics of the multiple possible future states,
as this is the only way to always solve the multiple choice problem, 
no matter what future state appears along with the distractors.
This common/shared representation is the invariance we are wishing for, \ie higher level semantics. 
In other words,
the representation of all these possible future states will be mapped to a space that their embeddings are close.


\section{Conclusion}
In this paper, we have introduced the Dense Predictive Coding~(DPC) framework for self-supervised representation learning on videos,
and outperformed the previous state-of-the-art by a large margin on the downstream tasks
of action classification on UCF101 and HMDB51.
As for future work, one straightforward extension of this idea is to employ different methods for aggregating the temporal
information -- instead of using a ConvGRU for temporal aggregation~($g(.)$ in the paper), 
other methods like masked CNN and attention based methods are also promising. 
In addition, empirical evidence shows that optical flow is able
to boost the performance for action recognition significantly;  it will
be interesting to explore how optical flow can be trained jointly with
DPC with self-supervised learning to further enhance the representation
quality.

\section*{Acknowledgements}
\vspace{-5pt}
Funding for this research is provided by the Oxford-Google DeepMind Graduate Scholarship, 
and by the EPSRC Programme Grant Seebibyte EP/M013774/1.

{\small
\bibliographystyle{ieee_fullname}
\bibliography{shortstrings,vgg_local,vgg_other}

\begin{thebibliography}{10}\itemsep=-1pt

\bibitem{Agrawal15}
Pulkit Agrawal, Joao Carreira, and Jitendra Malik.
\newblock Learning to see by moving.
\newblock In {\em ICCV}, 2015.

\bibitem{Anand19}
Ankesh Anand, Evan Racah, Sherjil Ozair, Yoshua Bengio, Marc{-}Alexandre
  C{\^{o}}t{\'{e}}, and R.~Devon Hjelm.
\newblock Unsupervised state representation learning in atari.
\newblock In {\em NIPS}, 2019.

\bibitem{Bengio03}
Yoshua Bengio, R{\'e}jean Ducharme, Pascal Vincent, and Christian Janvin.
\newblock A neural probabilistic language model.
\newblock In {\em JMLR}, 2003.

\bibitem{Caron18}
Mathilde Caron, Piotr Bojanowski, Armand Joulin, and Matthijs Douze.
\newblock Deep clustering for unsupervised learning of visual features.
\newblock In {\em ECCV}, 2018.

\bibitem{Doersch15}
Carl Doersch, Abhinav Gupta, and Alexei~A. Efros.
\newblock Unsupervised visual representation learning by context prediction.
\newblock In {\em CVPR}, 2015.

\bibitem{Feichtenhofer18}
Christoph Feichtenhofer, Haoqi Fan, Jitendra Malik, and Kaiming He.
\newblock Slowfast networks for video recognition.
\newblock {\em arXiv preprint arXiv:1812.03982}, 2018.

\bibitem{Fernando17}
Basura Fernando, Hakan Bilen, Efstratios Gavves, and Stephen Gould.
\newblock Self-supervised video representation learning with odd-one-out
  networks.
\newblock In {\em CVPR}, 2017.

\bibitem{Gidaris18}
Spyros Gidaris, Praveer Singh, and Nikos Komodakis.
\newblock Unsupervised representation learning by predicting image rotations.
\newblock In {\em ICLR}, 2018.

\bibitem{Gutmann10}
Michael~U. Gutmann and Aapo Hyv\"{a}rinen.
\newblock Noise-contrastive estimation: A new estimation principle for
  unnormalized statistical models.
\newblock In {\em AISTATS}, 2010.

\bibitem{Hara18}
Kensho Hara, Hirokatsu Kataoka, and Yutaka Satoh.
\newblock Can spatiotemporal 3d cnns retrace the history of 2d cnns and
  imagenet?
\newblock In {\em CVPR}, 2018.

\bibitem{Ioffe15}
Sergey Ioffe and Christian Szegedy.
\newblock Batch normalization: Accelerating deep network training by reducing
  internal covariate shift.
\newblock In {\em ICML}, 2015.

\bibitem{Isola15}
Phillip Isola, Daniel Zoran, Dilip Krishnan, and Edward~H Adelson.
\newblock Learning visual groups from co-occurrences in space and time.
\newblock In {\em ICLR}, 2015.

\bibitem{Jayaraman15}
Dinesh Jayaraman and Kristen Grauman.
\newblock Learning image representations tied to ego-motion.
\newblock In {\em ICCV}, 2015.

\bibitem{Jayaraman16}
Dinesh Jayaraman and Kristen Grauman.
\newblock Slow and steady feature analysis: higher order temporal coherence in
  video.
\newblock In {\em CVPR}, 2016.

\bibitem{Jing18}
Longlong Jing and Yingli Tian.
\newblock Self-supervised spatiotemporal feature learning by video geometric
  transformations.
\newblock {\em arXiv preprint arXiv:1811.11387}, 2018.

\bibitem{Kay17}
Will Kay, Jo{\~{a}}o Carreira, Karen Simonyan, Brian Zhang, Chloe Hillier,
  Sudheendra Vijayanarasimhan, Fabio Viola, Tim Green, Trevor Back, Paul
  Natsev, Mustafa Suleyman, and Andrew Zisserman.
\newblock The kinetics human action video dataset.
\newblock {\em arXiv preprint arXiv:1705.06950}, 2017.

\bibitem{Kim19}
Dahun Kim, Donghyeon Cho, and In~So Kweon.
\newblock Self-supervised video representation learning with space-time cubic
  puzzles.
\newblock In {\em AAAI}, 2019.

\bibitem{KingmaB14}
Diederik~P. Kingma and Jimmy Ba.
\newblock Adam: {A} method for stochastic optimization.
\newblock In {\em ICLR}, 2015.

\bibitem{Korbar18}
Bruno Korbar, Du Tran, and Lorenzo Torresani.
\newblock Cooperative learning of audio and video models from self-supervised
  synchronization.
\newblock In {\em NIPS}, 2018.

\bibitem{Kuehne11}
Hilde Kuehne, Huei-han Jhuang, Estibaliz Garrote, Tomaso Poggio, and Thomas
  Serre.
\newblock {HMDB}: A large video database for human motion recognition.
\newblock In {\em ICCV}, 2011.

\bibitem{Lai19}
Zihang Lai and Weidi Xie.
\newblock Self-supervised learning for video correspondence flow.
\newblock In {\em BMVC}, 2019.

\bibitem{Lee17}
Hsin{-}Ying Lee, Jia{-}Bin Huang, Maneesh Singh, and Ming{-}Hsuan Yang.
\newblock Unsupervised representation learning by sorting sequence.
\newblock In {\em ICCV}, 2017.

\bibitem{Lotter17}
William Lotter, Gabriel Kreiman, and David~D. Cox.
\newblock Deep predictive coding networks for video prediction and unsupervised
  learning.
\newblock In {\em ICLR}, 2017.

\bibitem{Mathieu16}
Michael Mathieu, Camille Couprie, and Yann LeCun.
\newblock Deep multi-scale video prediction beyond mean square error.
\newblock In {\em ICLR}, 2016.

\bibitem{Mikolov13a}
Tomas Mikolov, Kai Chen, Greg Corrado, and Jeffrey Dean.
\newblock Efficient estimation of word representations in vector space.
\newblock In {\em NIPS}, 2013.

\bibitem{Mikolov13b}
Tomas Mikolov, Ilya Sutskever, Kai Chen, Greg Corrado, and Jeffrey Dean.
\newblock Distributed representations of words and phrases and their
  compositionality.
\newblock In {\em NIPS}, 2013.

\bibitem{Misra16}
Ishan Misra, C.~Lawrence Zitnick, and Martial Hebert.
\newblock Shuffle and learn: Unsupervised learning using temporal order
  verification.
\newblock In {\em ECCV}, 2016.

\bibitem{Mnih13a}
Andriy Mnih and Koray Kavukcuoglu.
\newblock Learning word embeddings efficiently with noise-contrastive
  estimation.
\newblock In {\em NIPS}, 2013.

\bibitem{Noroozi16}
Mehdi Noroozi and Paolo Favaro.
\newblock Unsupervised learning of visual representations by solving jigsaw
  puzzles.
\newblock In {\em ECCV}, 2016.

\bibitem{Noroozi17}
Mehdi Noroozi, Hamed Pirsiavash, and Paolo Favaro.
\newblock Representation learning by learning to count.
\newblock In {\em ICCV}, 2017.

\bibitem{Pathak17}
Deepak Pathak, Ross~B. Girshick, Piotr Doll{\'{a}}r, Trevor Darrell, and
  Bharath Hariharan.
\newblock Learning features by watching objects move.
\newblock In {\em CVPR}, 2017.

\bibitem{Pathak16}
Deepak Pathak, Philipp Kr{\"{a}}henb{\"{u}}hl, Jeff Donahue, Trevor Darrell,
  and Alexei~A. Efros.
\newblock Context encoders: Feature learning by inpainting.
\newblock In {\em CVPR}, 2016.

\bibitem{Simonyan14b}
Karen Simonyan and Andrew Zisserman.
\newblock Two-stream convolutional networks for action recognition in videos.
\newblock In {\em NIPS}, 2014.

\bibitem{Soomro12}
Khurram Soomro, Amir~Roshan Zamir, and Mubarak Shah.
\newblock {UCF101}: A dataset of 101 human actions classes from videos in the
  wild.
\newblock {\em arXiv preprint arXiv:1212.0402}, 2012.

\bibitem{Srivastava14}
Nitish Srivastava, Geoffrey Hinton, Alex Krizhevsky, Ilya Sutskever, and Ruslan
  Salakhutdinov.
\newblock Dropout: A simple way to prevent neural networks from overfitting.
\newblock In {\em JMLR}, 2014.

\bibitem{Srivastava15}
Nitish Srivastava, Elman Mansimov, and Ruslan Salakhutdinov.
\newblock Unsupervised learning of video representations using {LSTM}s.
\newblock In {\em ICML}, 2015.

\bibitem{Sutskever14}
Ilya Sutskever, Oriol Vinyals, and Quoc~V. Le.
\newblock Sequence to sequence learning with neural networks.
\newblock In {\em NIPS}, 2014.

\bibitem{Tadmor16}
Oren Tadmor, Yonatan Wexler, Tal Rosenwein, Shai Shalev{-}Shwartz, and Amnon
  Shashua.
\newblock Learning a metric embedding for face recognition using the multibatch
  method.
\newblock In {\em NIPS}, 2016.

\bibitem{Ulyanov16}
Dmitry Ulyanov, Andrea Vedaldi, and Victor~S. Lempitsky.
\newblock Instance normalization: The missing ingredient for fast stylization.
\newblock {\em arXiv preprint arXiv:1607.08022}, 2016.

\bibitem{Oord18}
A{\"{a}}ron van~den Oord, Yazhe Li, and Oriol Vinyals.
\newblock Representation learning with contrastive predictive coding.
\newblock {\em arXiv preprint arXiv:1807.03748}, 2018.

\bibitem{Vondrick16b}
Carl Vondrick, Hamed Pirsiavash, and Antonio Torralba.
\newblock Anticipating visual representations from unlabelled video.
\newblock In {\em CVPR}, 2016.

\bibitem{Vondrick16a}
Carl Vondrick, Hamed Pirsiavash, and Antonio Torralba.
\newblock Generating videos with scene dynamics.
\newblock In {\em NIPS}, 2016.

\bibitem{Vondrick18}
Carl Vondrick, Abhinav Shrivastava, Alireza Fathi, Sergio Guadarrama, and Kevin
  Murphy.
\newblock Tracking emerges by colorizing videos.
\newblock In {\em ECCV}, 2018.

\bibitem{Wang15}
Xiaolong Wang and Abhinav Gupta.
\newblock Unsupervised learning of visual representations using videos.
\newblock In {\em ICCV}, 2015.

\bibitem{Wang19d}
Xiaolong Wang, Allan Jabri, and Alexei~A. Efros.
\newblock Learning correspondence from the cycle-consistency of time.
\newblock In {\em CVPR}, 2019.

\bibitem{Wei18}
Donglai Wei, Joseph Lim, Andrew Zisserman, and William~T. Freeman.
\newblock Learning and using the arrow of time.
\newblock In {\em CVPR}, 2018.

\bibitem{Wiskott02}
Laurenz Wiskott and Terrence Sejnowski.
\newblock Slow feature analysis: Unsupervised learning of invariances.
\newblock In {\em Neural Computation}, 2002.

\bibitem{Zhang16color}
Richard Zhang, Phillip Isola, and Alexei~A. Efros.
\newblock Colorful image colorization.
\newblock In {\em ECCV}, 2016.

\end{thebibliography}
}

\clearpage
\appendix
\noindent {\Large\textbf{Appendix}}\vspace{1mm}

\section{Architectures in detail}\label{sec:arch}
\begin{figure}[h!]
	\includegraphics[width=0.5\textwidth]{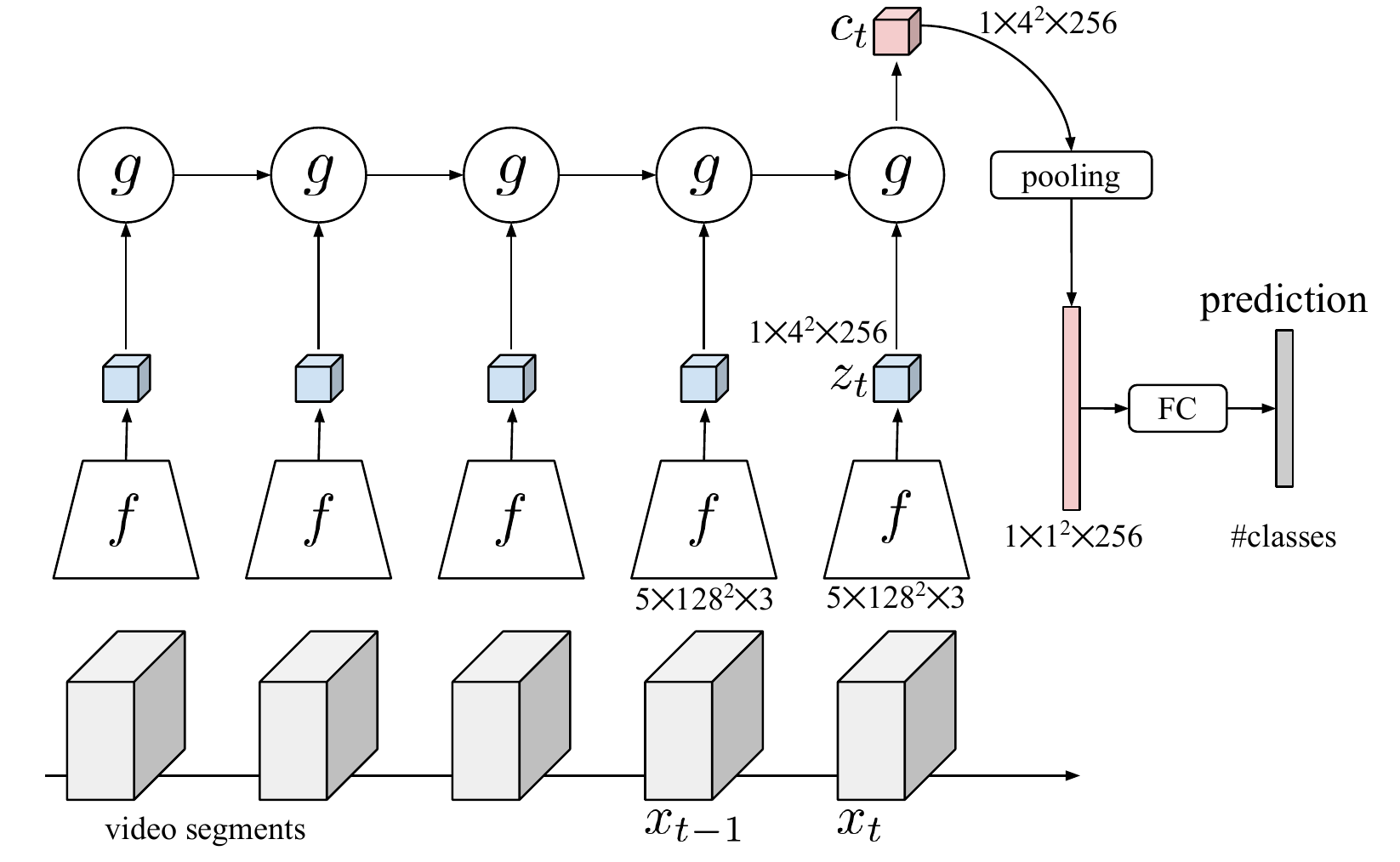}
	\centering
	\vspace{-2mm}
	\caption{The action classifier structure used to evaluate the representation.}\label{fig:arch_classifier}
\end{figure}

\vspace{-3mm}
We use tables to display CNN structures. 
The dimension of convolutional kernels are denoted by $\{ \text{temporal}\times\text{spatial}^2 \text{, channel size} \}$. 
The strides are denoted by $\{ \text{temporal stride, }\text{spatial stride}^2 \}$. 
The `output sizes' column displays the dimension of feature map after the operation (except the dimension of input data in the first row),
where $\{ t\times d^2 \times C \}$ denotes $\{ \text{temporal size}\times\text{spatial size}^2\times\text{channel size} \}$, 
and $T$ denotes the number of video blocks. In the following tables we take 3D-ResNet18 backbone with $128\times128$ input resolution as an example. 
\vspace{-3mm}

\paragraph{Structure of the action classifier.}
Table~\ref{table:arch_c} gives the details of the action
classifier which is used to evaluate the learned representation.
Figure~\ref{fig:arch_classifier} is a diagram of the action
classifier structure. For an input video with 30 fps, first a
temporal stride 3 is applied,
\ie every 3rd frame is taken, resulting in 10 fps. 
Then $T\times 5$ consecutive frames are sampled and truncated into $T$ video blocks,
\ie each video block has a size $5\times128^2\times3$, and we take $T=5$ for the action classifier. 

The action classifier is built with $f(.)$ and $g(.)$. 
The encoder function $f(.)$ takes 5 video blocks, each block contains 5 video frames ($5\times(5\times128^2\times3)$) as input,
spatio-temporal features~($z$) are extracted from the 5 video blocks with shared encoder~($f(.)$). 
Then the aggregation function $g(.)$~(ConvGRU) aggregates the 5 spatio-temporal feature maps into one spatio-temporal feature map, 
which is referred to as the context $c$ in the paper. 
The context $c$ is then pooled into a feature vector followed by a fully-connected layer. 

\vspace{-2mm}
\begin{table}[htbp!]
	\centering
	\begin{tabular}{c|c|c}
		\hline 
		module  & specification & $\begin{matrix}\text{output sizes}\\T\times t\times d^2\times C\end{matrix}$ \\ \hline 
		input data & - & $5\times(5\times128^2\times3)$\\ \hline 
		$f(.)$ & see Table~\ref{table:arch_f} & $5\times(1\times4^2\times256)$ ($z$) \\ \hline
		$g(.)$ & see Table~\ref{table:arch_g} & $1\times1\times4^2\times256$ ($c$) \\ \hline
		pool & $\begin{matrix}1\times4^2\\\text{stride }1,1^2\end{matrix}$ & $1\times1\times1^2\times256$ \\ \hline
		final fc & 1-layer FC & $1\times1\times1^2\times\text{\# classes}$ \\ \hline
		& \multicolumn{2}{c}{compute cross-entropy loss} \\
		\hline 
	\end{tabular}
	\vspace{-2mm}
	\caption{The structure of the linear classifier.}\label{table:arch_c}
	\vspace{-3mm}
\end{table}

\paragraph{Structure of the DPC.}

The DPC is built from $f(.)$ and $g(.)$ with an additional prediction
mechanism, which is described in Table~\ref{table:arch_dpc}. Here we use 5pred3 setting for an example, where $f(.)$ takes 5 video blocks and
extracts 5 spatio-temporal feature maps, then $g(.)$ aggregates feature
maps into context $c$. The prediction function $\phi(.)$ is a two-layer
perceptron, which takes the context $c$ as input and produces a
predicted feature $\hat{z}$ as output. The contrastive loss is
computed using $z$ and $\hat{z}$ as described in the paper Sec.~\ref{sec:method-loss}.

\vspace{-2mm}
\begin{table}[htbp!]
	\centering
	\begin{tabular}{c|c|c}
		\hline 
		module  & specification & $\begin{matrix}\text{output sizes}\\T\times t\times d^2\times C\end{matrix}$ \\ \hline 
		input data & - & $5\times(5\times128^2\times3)$\\ \hline 
		$f(.)$ & see Table~\ref{table:arch_f} & $5\times(1\times4^2\times256)$ ($z$) \\ \hline
		$g(.)$ & see Table~\ref{table:arch_g} & $1\times1\times4^2\times256$ ($c$) \\ \hline
		$\phi(.)$ & 2-layer FC & $1\times1\times4^2\times256$ ($\hat{z}$) \\ \hline
		& \multicolumn{2}{c}{compute loss using $z$ and $\hat{z}$} \\
		\hline 
	\end{tabular}
	\vspace{-2mm}
	\caption{The structure of DPC model. }\label{table:arch_dpc}
	\vspace{-3mm}
\end{table}

\vspace{-2mm}
\paragraph{Structure of $f(.)$.}

The detailed structure of the encoder function $f(.)$ is shown in Table~\ref{table:arch_f}. Note that $f(.)$ takes input video blocks independently, so the number of video block $T$ is omitted in the table. 

\vspace{-2mm}
\begin{table}[htbp!]
	\centering
	\begin{tabular}{c|c|c}
		\hline 
		stage   &  specification  & $ \begin{matrix} \text{output sizes}\\ t\times d^2\times C \end{matrix}$ \\ \hline 
		input data & - & $5\times 128^2\times 3$ \\ \hline 
		$\text{conv}_1$  & $\begin{matrix} 1\times 7^{2}, 64\\ \text{stride } 1, 2^2\end{matrix}$ & $5\times64^2\times64$ \\ \hline 
		$\text{pool}_1$    & $\begin{matrix} 1\times 3^{2}, 64\\ \text{stride } 1, 2^2\end{matrix}$ & $5\times32^2\times64$ \\ \hline 
		$\text{res}_2$     & $\begin{bmatrix} 1\times3^2, 64\\1\times3^2, 64 \end{bmatrix}\times 2$ & $5\times32^2\times64$ \\ \hline 
		$\text{res}_3$     & $\begin{bmatrix} 1\times3^2, 128\\1\times3^2, 128 \end{bmatrix}\times 2$ & $5\times16^2\times128$ \\ \hline 
		$\text{res}_4$     & $\begin{bmatrix} 3\times3^2, 256\\3\times3^2, 256 \end{bmatrix}\times 2$ & $3\times8^2\times256$ \\ \hline 
		$\text{res}_5$     & $\begin{bmatrix} 3\times3^2, 256\\3\times3^2, 256 \end{bmatrix}\times 2$ & $2\times4^2\times256$ \\ \hline 
		$\text{pool}_2$    & $\begin{matrix} 2\times 1^{2}, 256\\ \text{stride } 1, 1^2\end{matrix}$ & $1\times4^2\times256$ \\ 
		\hline 
	\end{tabular}
	\vspace{-2mm}
	\caption{The structure of the encoding function $f(.)$ with 3D-ResNet18 backbone. }\label{table:arch_f}
	\vspace{-3mm}
\end{table}

\vspace{-3mm}
\paragraph{Structure of $g(.)$.}

The structure of the temporal aggregation function $g(.)$ is shown in Table~\ref{table:arch_g}. It aggregates the feature maps over the past $T$ time steps. Note that in the case of sequential prediction, $T$ increments by 1 after each prediction step. Table~\ref{table:arch_g} shows the case where $g(.)$ aggregates the feature maps over the past 5 steps. 

\vspace{-2mm}
\begin{table}[htbp!]
	\centering
	\begin{tabular}{c|c|c}
		\hline 
		stage  & specification & $\begin{matrix}\text{output sizes}\\T\times t\times d^2\times C\end{matrix}$ \\ \hline 
		input data & - & $5\times1\times4^2\times256$\\ \hline 
		ConvGRU &$\begin{matrix} [1^2, 256]\end{matrix}\times1 \text{ layer}$ & $1\times1\times4^2\times256$ \\
		\hline  
	\end{tabular}
	\vspace{-2mm}
	\caption{The structure of aggregation function $g(.)$. }\label{table:arch_g}
	\vspace{-3mm}
\end{table}

\section{t-SNE clustering of DPC context representation}\label{appendix:t-sne}

\begin{figure}[h!]
	\includegraphics[width=0.5\textwidth]{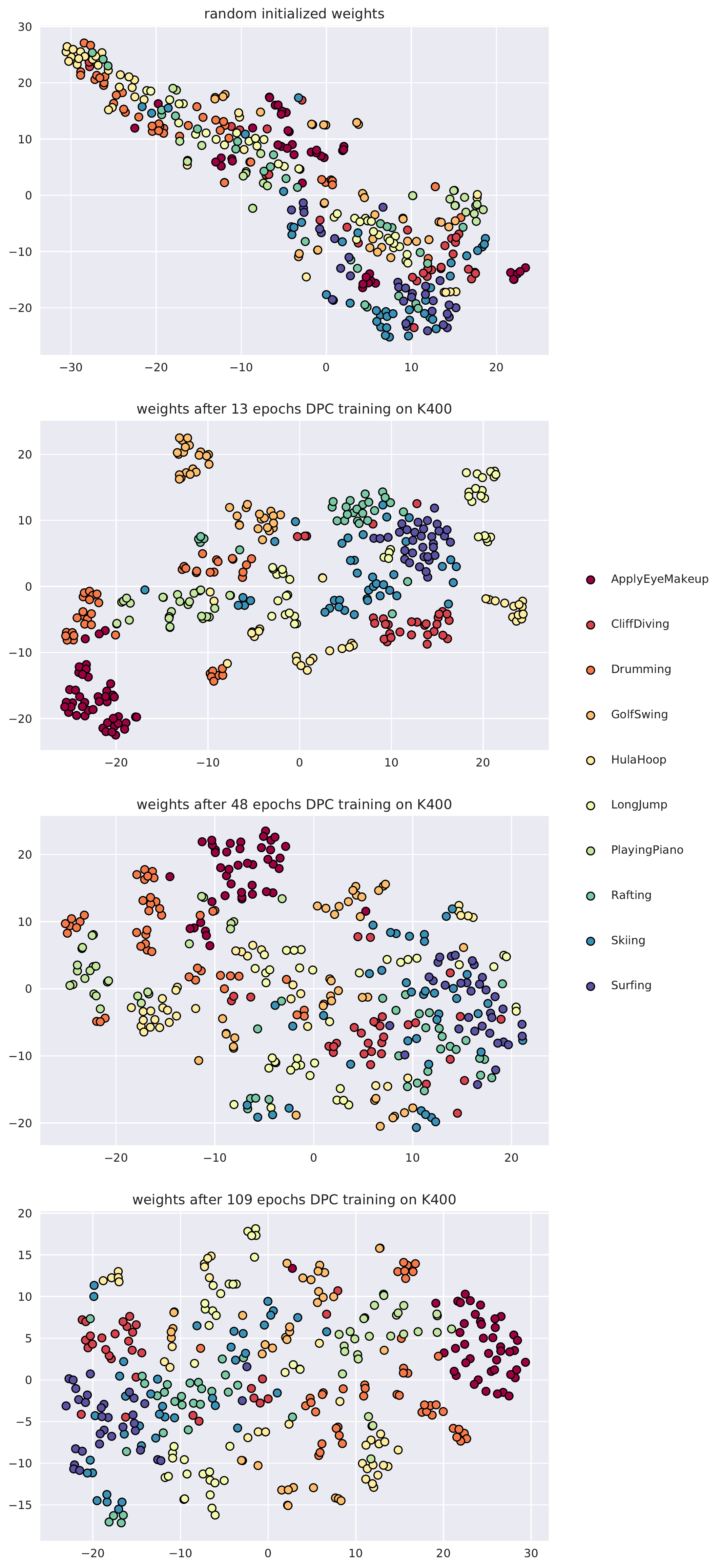}
	\centering
	\vspace{-5mm}
	\caption{t-SNE visualization of the context representations on UCF101 validation set extracted by different $f(.)$ and $g(.)$ after $\{0 \text{ (random init.)}, 13, 48, 109\}$ epochs DPC training on K400.}\label{fig:tsne}
	\vspace{-3mm}
\end{figure}

This section shows the t-SNE clustering of the context representation on UCF101 extracted by $f(.)$ and $g(.)$ (Figure~\ref{fig:tsne}). 
In detail, 5 consecutive video blocks are sampled from each video in the validation set, then the feature maps $\{z_1, ..., z_5\}$ are extracted from each video block and aggregated into context representation $c_5$ and then pooled into vectors. We use t-SNE to visualize the context vectors in 2D. 
For clarity, only 10 action classes (out of 101 classes from UCF101) are displayed. 
The upper-left figure visualizes the context features extracted by randomly initialized $f(.)$ and $g(.)$. 
The following 3 figures show the context features extracted by $f(.)$ and $g(.)$ 
after $\{13, 48, 109\}$ epochs of DPC training on K400, without any finetuning on UCF101. 

It can be seen that as the DPC training proceeds the intra-class distance is reduced (compared to the random initialization) 
and also the inter-class distance is increased,~\ie the self-supervised DPC method 
is clustering the feature vectors into action classes.

\section{Cosine distance histogram of DPC context representation}\label{appendix:cos}

This section shows the cosine distance of the context representation on UCF101 extracted by DPC pre-trained $f(.)$ and $g(.)$ (Figure~\ref{fig:cos}).
We use the same setting as Figure~\ref{fig:tsne} and extract one context representation for each video and pool into vector. Then we compute the cosine distance of each pair of context vectors across the entire UCF101 validation set. The cosine distance is summarized by histogram, where `positive' means two source videos are from the same action class and `negative' means two source videos are from different action classes. For clarity, 17 out of 101 action classes are evenly sampled from UCF101 and visualized. Note that there is no finetunning in this stage,~\ie the network doesn't see any action labels. 

It can be seen that for all action classes, the context representations from the same action class have higher cosine similarity,~\ie DPC can cluster actions without knowing action labels. 

\twocolumn[{
	\centering
	\includegraphics[width=1\textwidth]{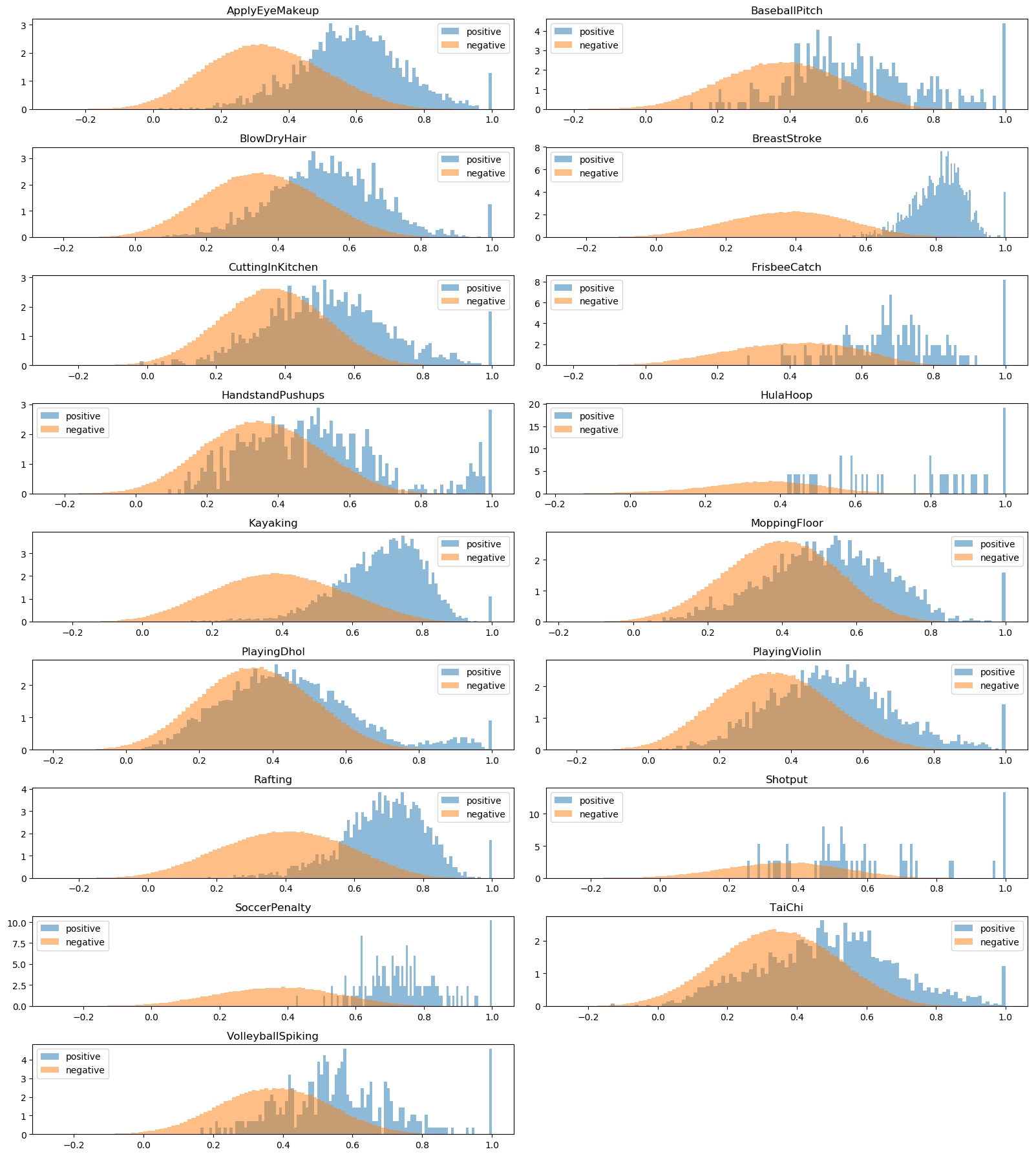}
	\captionof{figure}{Histogram of the cosine distance of the context representations extracted from UCF101 validation set by DPC weights. `Positive' and `negative' refer to the video pairs that are from the same or different action classes. DPC is trained on K400 without any finetuning on UCF101.}\label{fig:cos}
}]


\end{document}